\documentclass[10pt]{article} 
\usepackage{bm}
\usepackage{amsmath}
\usepackage[]{algorithmic} 
\usepackage[boxed]{algorithm}
\usepackage{graphicx}
\usepackage{pifont}
\usepackage{amssymb}
\usepackage{color}
\include{header}
\usepackage{courier}

\newcommand*\pd[3][]
{ 
\def\next{#1}%
\ifx\empty\next
  \frac{\partial#2}{\partial #3}
\else
  \frac{{\partial^{#1} #2}}{\partial#3^{#1}}
\fi
}   
\title{\textbf{Stochastic Backpropagation through\\Mixture Density Distributions}}
\date{}
\author{Alex Graves\\\small{Google DeepMind, London, UK}\\\small{\texttt{gravesa@google.com}}}
\begin{document} 
\maketitle
\begin{abstract}
The ability to backpropagate stochastic gradients through continuous latent distributions has been crucial to the emergence of variational autoencoders~\cite{gregor2013deep,kingma2013auto,rezende2014stochastic,gregor15draw} and stochastic gradient variational Bayes~\cite{graves11nips,kingma2015variational,blundell15uncertainty}.
The key ingredient is an unbiased and low-variance way of estimating gradients with respect to distribution parameters from gradients evaluated at distribution samples.
The ``reparameterization trick''~\cite{kingma2013auto} provides a class of transforms yielding such estimators for many continuous distributions, including the Gaussian and other members of the location-scale family.
However the trick does not readily extend to mixture density models, due to the difficulty of reparameterizing the discrete distribution over mixture weights.
This report describes an alternative transform, applicable to any continuous multivariate distribution with a differentiable density function from which samples can be drawn, and uses it to derive an unbiased estimator for mixture density weight derivatives.
Combined with the reparameterization trick applied to the individual mixture components, this estimator makes it straightforward to train variational autoencoders with mixture-distributed latent variables, or to perform stochastic variational inference with a mixture density variational posterior.
\end{abstract}
\section*{General Result}
Let $f(\mathbf{x})$ be a probability density function (PDF) over $\mathbf{x} \in \mathbb{R}^D$ and cumulative density function (CDF) $F(\mathbf{x})$.
$f$ can be rewritten as
\begin{equation}
f(\mathbf{x}) = \prod_{d=1}^Df_d(x_d|\mathbf{x}_{<d})
\end{equation}
where $\mathbf{x}_{<d} = x_1,\dots,x_{d-1}$, $f_1(x_1|\mathbf{x}_{<1}) = f_1(x_1)$ and 
$f_d$ is the marginal PDF of $x_d$ conditioned on $\mathbf{x}_{<d}$.
A sample $\hat{\mathbf{x}}$ can be drawn from $f$ using the multivariate quantile transform: first draw a vector of $D$ independent samples $\mathbf{u} = (u_1,\dots,u_D)$ from $U(0,1)$, then recursively define $\hat{\mathbf{x}}$ as
\begin{align}
\hat{x}_1 &= F_1^{-1}(u_1)\\
\label{eq:inverse}
\hat{x}_d &= F_d^{-1}(u_d|\hat{\mathbf{x}}_{<d})
\end{align}
where $F_d^{-1}$ is the quantile function (inverse CDF) corresponding to the PDF $f_d$. 
Inverting Eq.~\ref{eq:inverse} and applying the definition of a univariate CDF yields
\begin{equation}
F_d(\hat{x}_d|\hat{\mathbf{x}}_{<d}) = \int_{t=-\infty}^{\hat{x}_d}{f_d(t|\hat{\mathbf{x}}_{<d})dt} = u_d
\end{equation}
Assume that $f$ depends on some parameter $\theta$.
The general form of Leibniz integral rule tells us that 
\begin{equation}
\pd{F_d(\hat{x}_d|\hat{\mathbf{x}}_{<d})}{\theta} = f_d(\hat{x}_d|\hat{\mathbf{x}}_{<d})\pd{\hat{x}_d}{\theta} + \int_{t=-\infty}^{\hat{x}_d}{\pd{f_d(t|\hat{\mathbf{x}}_{<d})}{\theta}dt} = \pd{u_d}{\theta} = 0
\end{equation}
and therefore
\begin{equation}
\label{eq:dx_dtheta}    
\pd{\hat{x}_d}{\theta} = -\frac{1}{f_d(\hat{x}_d|\hat{\mathbf{x}}_{<d})} \int_{t=-\infty}^{\hat{x}_d}{\pd{f_d(t|\hat{\mathbf{x}}_{<d})}{\theta}dt}    
\end{equation}
If the above integral is intractable it can be estimated with Monte-Carlo sampling, as long as $f_d(t |\hat{\mathbf{x}}_{<d})$ can be sampled and $F_d(\hat{x}_d|\hat{\mathbf{x}}_{<d})$ is tractable:
\begin{align}
&\int_{t=-\infty}^{\hat{x}_d}{\pd{f_d(t|\hat{\mathbf{x}}_{<d})}{\theta}dt} = \int_{t=-\infty}^{\hat{x}_d}{f_d(t|\hat{\mathbf{x}}_{<d}) \pd{\log f_d(t|\hat{\mathbf{x}}_{<d})}{\theta}dt}\\
&=F_d(\hat{x}_d|\hat{\mathbf{x}}_{<d}) \int_{t=-\infty}^{\infty}{f_d(t \leq \hat{x}_d|\hat{\mathbf{x}}_{<d}) \pd{\log f_d(t|\hat{\mathbf{x}}_{<d})}{\theta}dt}\\
&\approx  \frac{F_d(\hat{x}_d|\hat{\mathbf{x}}_{<d})}{N}\sum_{n=1}^N \pd{\log f_d(t^n|\hat{\mathbf{x}}_{<d})}{\theta};\ t^n \sim f_d(t \leq \hat{x}_d|\hat{\mathbf{x}}_{<d})
\end{align}
where 
\begin{equation}
f_d(t \leq \hat{x}_d|\hat{\mathbf{x}}_{<d}) = \begin{cases} \frac{f_d(t|\hat{\mathbf{x}}_{<d})}{F_d(\hat{x}_d|\hat{\mathbf{x}}_{<d})}\text{ if } t \leq \hat{x}_d\\
0 \text{ otherwise}\end{cases}
\end{equation}
which can be sampled by drawing from 
$f_d(t |\hat{\mathbf{x}}_{<d})$ and rejecting the result if it is greater than $\hat{x}_d$.

Let $h$ be the expectation over $f$ of an arbitrary differentiable function $g$ of $\mathbf{x}$ (e.g. a loss function) and denote by $Q(\mathbf{u})$ the sample from $f$ returned by the quantile transform applied to $\mathbf{u}$. Then
\begin{equation}
h = \int_{\mathbf{u}\in [0, 1]^D}{{g(Q(\mathbf{u}))d\mathbf{u}}}
\end{equation}
and hence
\begin{align}
\pd{h}{\theta} &= \int_{\mathbf{u} \in [0,1]^D}{\pd{{g(Q(\mathbf{u}))}}{\theta}d\mathbf{u}}\\
 &= \int_{\mathbf{u} \in [0, 1]^D}\sum_{d=1}^D\pd{{g(Q(\mathbf{u}))}}{Q_d(\mathbf{u})}\pd{Q_d(\mathbf{u})}{\theta} d\mathbf{u}
\end{align}
which can be estimated with Monte-Carlo sampling:
\begin{align}
\label{eq:mc_loss_deriv}
\pd{h}{\theta} &\approx \frac{1}{N}\sum_{n=1}^N{\sum_{d=1}^D\pd{g(\mathbf{x}^n)}{x^n_d}  \pd{x_d^n}{\theta}}
\end{align}
where $\mathbf{x}^n \sim f(\mathbf{x})$. 
Note that the above estimator does not require $Q$ to be known, as long as $f$ can be sampled. 

\section*{Application to Mixture Density Weights}
If $f$ is a mixture density distribution with $K$ components then
\begin{equation}
f(\mathbf{x}) = \sum_{k=1}^K{\pi_k f^k(\mathbf{x})}
\end{equation}
and
\begin{equation}
\label{eq:mixture_density}
f_d(x_d|\mathbf{x}_{<d}) = \sum_{k=1}^K{\Pr(k|\mathbf{x}_{<d})f^k_d(x_d|\mathbf{x}_{<d})}
\end{equation}
where $\Pr(k|\mathbf{x}_{<d})$ is the posterior responsibility of the component $k$, given the prior mixture density weight $\pi_k$ and the observation sequence $\mathbf{x}_{<d}$. 

In what follows we will assume that the mixture components have diagonal covariance, so that $f^k_d(\mathbf{x}_{d}|\mathbf{x}_{<d}) = f^k_d(\mathbf{x}_{d})$. It should be possible to extend the analysis to non-diagonal components, but that is left for future work.
Abbreviating $\Pr(k|\mathbf{x}_{<d})$ to $p^k_d$ and applying the diagonal covariance of the components, Eq.~\ref{eq:mixture_density} becomes
\begin{equation}
\label{eq:mixture_density_abbrv}
f_d(x_d|\mathbf{x}_{<d}) = \sum_{k}{p^k_d f^k_d(x_d)}
\end{equation}
where $p^k_d$ is defined by the following recurrence relation:
\begin{align}
\label{eq:pk1}
p^k_1 &= \pi_k\\
\label{eq:mixture_density_prob_recursion}
p^k_d &= \frac{p^k_{d-1} f^k_{d-1}(x_{d-1})}{f_{d-1}(x_{d-1}|\mathbf{x}_{<d-1})}
\end{align}
We seek the derivatives of $h$ with respect to the mixture weights $\pi_j$, after the weights have been normalised (e.g.\ by a softmax function).
Setting $x_d = t$ and differentiating Eq.~\ref{eq:mixture_density_abbrv} gives
\begin{align}
\pd{f_d(t|\mathbf{x}_{<d})}{\pi_j} &= \sum_k\left[ \pd{p^k_d}{\pi_j}f^k_d(t) + \pd{f^k_d(t)}{t}\pd{t}{\pi_j}p^k_d \right]
\end{align}
Setting $\mathbf{x} = \hat{\mathbf{x}}$ where $\hat{\mathbf{x}}$ is a sample drawn from $f$, and observing that $\pd{t}{\pi_j} = 0$ if $t$ does not depend on $f$, we can substitute the above into Eq.~\ref{eq:dx_dtheta} to get
\begin{align}
\pd{\hat{x}_d}{\pi_j} &= -\frac{1}{f_d(\hat{x}_d|\hat{\mathbf{x}}_{<d})}\sum_k\pd{p^k_d}{\pi_j} \int_{t=-\infty}^{\hat{x}_d} f^k_d(t) dt\\
&= -\frac{1}{f_d(\hat{x}_d|\hat{\mathbf{x}}_{<d})}\sum_k\pd{\log p^k_d}{\pi_j} p^k_d F^k_d(\hat{x}_d)
\end{align}
Differentiating Eq.~\ref{eq:mixture_density_prob_recursion} yields (after some rearrangement)
\begin{align}
\pd{\log p^k_d}{\pi_j} &= \pd{\log p^k_{d-1}}{\pi_j} - \sum_l p^l_d \pd{\log p^l_{d-1}}{\pi_j}\\&+ \left[\pd{\log f^k_{d-1}(\hat{x}_{d-1})}{\hat{x}_{d-1}} - \sum_l p^l_d \pd{\log f^l_{d-1}(\hat{x}_{d-1})}{\hat{x}_{d-1}}\right] \pd{\hat{x}_{d-1}}{\pi_j}
\end{align}
$\pd{\log p^k_d}{\pi_j}$ and $\pd{\hat{x}_d}{\pi_j}$ can then be obtained with a joint recursion, starting from the initial conditions
\begin{align}
\pd{\log p^k_1}{\pi_j} &= \frac{\delta_{jk}}{\pi_j}\\
\pd{\hat{x}_1}{\pi_j} &= -\frac{F_1^j(\hat{x}_1)}{f_1(\hat{x}_1)}
\end{align}
We are now ready to approximate $\pd{h}{\pi_j}$ by substituting into Eq.~\ref{eq:mc_loss_deriv}:
\begin{equation}
\pd{h}{\pi_j} \approx \frac{1}{N}\sum_{n=1}^N\sum_{d=1}^D\pd{g(\mathbf{x}^n)}{x^n_d}\pd{x^n_d}{\pi_j};\ \mathbf{x}^n \sim f(\mathbf{x})
\end{equation}
Pseudocode for the complete computation is provided in Algorithm~\ref{fig:pseudocode}.

\begin{algorithm}[H]
\caption{Stochastic Backpropagation through Mixture Density Weights}
\begin{algorithmic}
\STATE initialise $\pd{h}{\pi_j} \leftarrow 0$
\FOR{$n=1$ to $N$}
\STATE draw $\mathbf{x} \sim f(\mathbf{x})$
\STATE $p^k_1 \leftarrow \pi_k$
\STATE $\pd{\log p^k_1}{\pi_j} \leftarrow \frac{\delta_{jk}}{\pi_j}$
\STATE $\pd{x_1}{\pi_j} \leftarrow -\frac{F^j_1(x_1)}{f_1(x_1)}$
\STATE $f_1(x_1) \leftarrow \sum_k \pi_k f^k_1(x_1)$
\FOR{$d=2$ to $D$}
\STATE $f_d(x_d|\mathbf{x}_{<d}) \leftarrow \sum_k p^k_d f^k_d(x_d)$
\STATE $p^k_d \leftarrow \frac{p^k_{d-1} f^k_{d-1}(x_{d-1})}{f_{d-1}(x_{d-1}|\mathbf{x}_{<d-1})}$
\STATE $\pd{\log p^k_d}{\pi_j} \leftarrow \pd{\log p^k_{d-1}}{\pi_j} - \sum_l p^l_d \pd{\log p^l_{d-1}}{\pi_j} +$ \par
\hskip4.5\algorithmicindent $\pd{x_{d-1}}{\pi_j}\left[\pd{\log f^k_{d-1}(x_{d-1})}{x_{d-1}} - \sum_l p^l_d \pd{\log f^l_{d-1}(x_{d-1})}{x_{d-1}}\right]$
\STATE $\pd{x_d}{\pi_j} \leftarrow -\frac{1}{f_d(x_d|\mathbf{x}_{<d})}\sum_k \pd{\log p^k_d}{\pi_j} p^k_d F^k_d(x_d)$
\ENDFOR
\STATE $\pd{h}{\pi_j} \leftarrow \pd{h}{\pi_j} + \sum_d \pd{g(\mathbf{x})}{x_d} \pd{x_d}{\pi_j}$ 
\ENDFOR
\STATE $\pd{h}{\pi_j} \leftarrow \frac{1}{N}\pd{h}{\pi_j}$
\end{algorithmic}
\label{fig:pseudocode}
\end{algorithm}

\section*{Acknowledgements}
Useful discussions and comments were provided by Ivo Danihelka, Danilo Rezende, Remi Munos, Diederik Kingma, Charles Blundell, Mevlana Gemici, Nando de Freitas, and Andriy Mnih.

\bibliography{mixture_derivatives}

\begin{thebibliography}{1}

\bibitem{blundell15uncertainty}
C.~{Blundell}, J.~{Cornebise}, K.~{Kavukcuoglu}, and D.~{Wierstra}.
\newblock {Weight Uncertainty in Neural Networks}.
\newblock {\em ArXiv e-prints}, May 2015.

\bibitem{graves11nips}
A.~Graves.
\newblock Practical variational inference for neural networks.
\newblock In {\em Advances in Neural Information Processing Systems},
  volume~24, pages 2348--2356. 2011.

\bibitem{gregor15draw}
K.~Gregor, I.~Danihelka, A.~Graves, and D.~Wierstra.
\newblock {DRAW:} {A} recurrent neural network for image generation.
\newblock {\em ArXiv e-prints}, March 2015.

\bibitem{gregor2013deep}
K.~Gregor, I.~Danihelka, A.~Mnih, C.~Blundell, and D.~Wierstra.
\newblock Deep autoregressive networks.
\newblock In {\em Proceedings of the 31st International Conference on Machine
  Learning}, 2014.

\bibitem{kingma2015variational}
D.~P. Kingma, T.~Salimans, and M.~Welling.
\newblock Variational dropout and the local reparameterization trick.
\newblock {\em ArXiv e-prints}, June 2015.

\bibitem{kingma2013auto}
D.~P. Kingma and M.~Welling.
\newblock Auto-encoding variational bayes.
\newblock In {\em Proceedings of the International Conference on Learning
  Representations}, 2014.

\bibitem{rezende2014stochastic}
D.~J. Rezende, S.~Mohamed, and D.~Wierstra.
\newblock Stochastic backpropagation and approximate inference in deep
  generative models.
\newblock In {\em Proceedings of the 31st International Conference on Machine
  Learning}, pages 1278--1286, 2014.

\end{thebibliography}
\bibliographystyle{abbrv}
\end{document}